\documentclass[a4paper]{article}

\usepackage{INTERSPEECH_v2}

\usepackage{array}
\newcolumntype{B}{>{\centering\arraybackslash}p{0.8cm}}
\newcolumntype{C}{>{\centering\arraybackslash}p{1.1cm}}
\newcolumntype{X}[1]{>{\centering\arraybackslash}p{#1}}
\newcolumntype{E}{>{\arraybackslash}p{0.7cm}}

\title{A Batch Noise Contrastive Estimation Approach for Training Large Vocabulary Language Models}
\name{Youssef Oualil, Dietrich Klakow}
\address{
  Spoken Language Systems (LSV) \\
  Collaborative Research Center on Information Density and Linguistic Encoding \\
  Saarland University, Saarbr\"{u}cken, Germany
	}
\email{\{firstname.lastname\}@lsv.uni-saarland.de}

\begin{document}
\maketitle
%
\begin{abstract}
\label{sec:abstract}
%
Training large vocabulary Neural Network Language Models (NNLMs) is a difficult
task due to the explicit requirement of the output layer normalization, which
typically involves the evaluation of the full softmax function over the complete vocabulary.
This paper proposes a Batch Noise Contrastive Estimation (B-NCE) approach to alleviate this
problem. This is achieved by reducing the vocabulary, at each time step, to
the target words in the batch and then replacing the softmax by the noise
contrastive estimation approach, where these words play the role of targets and
noise samples at the same time. In doing so, the proposed approach can be fully
formulated and implemented using optimal dense matrix operations. Applying
B-NCE to train different NNLMs on the Large Text Compression Benchmark (LTCB)
and the One Billion Word Benchmark (OBWB) shows a significant reduction of the
training time with no noticeable degradation of the models performance. This
paper also presents a new baseline comparative study of different standard 
NNLMs on the large OBWB on a single Titan-X GPU.
\end{abstract}


\section{Introduction}
\label{sec:intro}

Neural Network Language Models (NNLM)~\cite{Bengio2003a,Mikolov2010} have been shown to significantly outperform standard N-gram LMs~\cite{Rosenfeld2000,KN1995}
on many speech and language technology applications, such as machine translation~\cite{Bahdanau2015} and speech recognition~\cite{Hinton2012}. 
The training and evaluation of these models, however, becomes significantly slow and challenging when considering 
large vocabulary language models~\cite{Chelba2013} . This is mainly due to the explicit 
normalization of the output layer, which typically requires the evaluation of the full \textit{softmax} 
function over the complete vocabulary. 

In order to overcome this problem, Schwenk et al.~\cite{Schwenk2005} proposed to use a short list of frequent words 
in combination with N-gram models. The performance of this approach, however, significantly depends on the short list size. 
In a different attempt, Morin et al. ~\cite{Morin2005} proposed to factorize the output probabilities using a binary tree, 
which results in an exponential speed-up of the training and evaluation, whereas Mikolov et al.~\cite{Mikolov2011} proposed
to use an additional class layer as an alternative to the factorization. The performance of these two approaches significantly depends
on the design of the binary tree and the class layer size, respectively. 
As an alternative to modifying the architecture design, the authors of~\cite{Bengio2003b} used importance sampling to approximate the output gradient.
Unfortunately, this approach requires a control of the samples variance, which can lead otherwise to unstable learning~\cite{Bengio2008}. 
In a similar work, Mnih et al.~\cite{Mnih2012} proposed to use Noise Contrastive Estimation (NCE)~\cite{Gutmann2012} to speed-up 
the training. NCE treats the learning as a binary classification problem between a target word and noise samples, which are drawn from 
a noise distribution. Moreover, NCE considers the normalization term as an additional parameter that can be learned during training or fixed beforehand. In this case, 
the network learns to \textit{self-normalize}. This property makes NCE more attractive compared to other sampling methods, such as importance sampling, 
which would still require the use of the softmax function during evaluation.
In batch mode training, however, the implementation of NCE cannot be directly formulated using dense matrix operations, which compromises their speed-up gains. 

This paper proposes a new solution to train large vocabulary LMs using NCE in batch mode. The main idea here is to restrict the vocabulary, 
at each iteration, to the words in the batch and replace the standard \textit{softmax} function by NCE. In particular, 
the target words (to predict) in the batch play the role of targets and noise samples at the same time. In doing so, 
the proposed approach does not require any sampling and can be fully formulated using dense matrix operations, 
leading to significant speed-up improvements with no noticeable degradation of the performance. Moreover, we 
can show that this approach optimally approximates the unigram noise distribution, which is widely used in NCE-based 
LMs (Section~\ref{sec:bnce}). 

While applying the proposed batch NCE approach, this paper also presents a new baseline comparative study 
of different NNLMs on the Large Text Compression Benchmark (LTCB)~\cite{Mahoney2011} and the 
One Billion Word Benchmark (OBWB)~\cite{Chelba2013} on a \textbf{single Titan-X GPU} (Section~\ref{sec:exp}).

\section{Noise Contrastive Estimation}
\label{sec:nce}
Probabilistic NNLM generally require the normalization of the output layer to produce meaningful probability
distributions. Using the \textit{softmax} function, the probability of the next word $w$, given the context $c$ and the model parameters $\theta$, 
is given by 
%
\begin{equation}
\label{eq:softmax}
\displaystyle{ p_{\theta}^c(w) = p(w|c,\theta) = \frac{exp(\mathcal{M}_{\theta}(w,c))}{\sum_{v\in V}{exp(\mathcal{M}_{\theta}(v,c))}} }   
\end{equation}
$V$ is the vocabulary and $\mathcal{M}_\theta$ is the neural scoring function.
Learning the parameters $\theta$ generally requires the evaluation of the 
normalization term for each context $c$. This evaluation involves the complete vocabulary 
and therefore, becomes very challenging and resource demanding for large vocabulary corpora.
As an alternative solution,~\cite{Gutmann2012} proposed to use noise contrastive estimation to train unnormalized probabilistic models.
Thus, $p_{\theta}^c(w)$ is approximated as
%
\begin{equation}
\label{eq:prob}
  \displaystyle{ p_{\theta}^c(w) \approx \frac{exp(\mathcal{M}_{\theta}(w,c))}{ Z_c }} 
\end{equation}
$Z_c$ is a context-dependent normalization term, which is fixed in practice to a constant value $Z$ (e.g., $Z=1$ in~\cite{Zoph2016} and $Z=exp(9)$ in~\cite{Chen2015}. 
The latter constant will be used in the experiments conducted in Section~\ref{sec:exp}).

The idea behind NCE is to cast the density estimation problem to a binary classification task, which learns to discriminate between
real samples drawn from the data distribution $p_D^c$ and noise samples drawn from a given noise distribution $p_n^c$. 
Although $p_n^c$ is context-dependent, it has been shown (e.g.~\cite{Mnih2012}) that context-independent noise distributions such as unigram, are sufficient  
to achieve a good performance. Thus, $p_n^c=p_n$ in the rest of this paper. If $K$ denotes the number of noise samples,
the probability that the word $w$ is generated from the data distribution ($L=1$) or noise distribution ($L=0$) are given by 
%
\vspace{1mm}
\begin{align}
  \label{eqn:true}
    p_c^w(1)   \overset{\Delta}{=}  p(L=1|w,c)  &=  \frac{p_{\theta}^c(w)}{p_{\theta}^c(w) + K \cdot p_n(w)} \\
   \label{eqn:false}
    p_c^w(0)   \overset{\Delta}{=}  p(L=0|w,c)  &=  1 - p(L=1|w,c)
 \end{align}

According to NCE, the model distribution $p_{\theta}^c$ is expected to converge towards the data distribution $p_D^c$ after minimizing the 
following objective function on the data $D$
%
%
\begin{equation}
\label{eq:llog}
\displaystyle{ \mathcal{J}(\theta) = -\sum_{w_i}^D \! {\left( log(p_{c_i}^{w_i}(1))  + \sum_{k}{log(p_{c_i}^{w_i^k}(0))} \right) }} 
\end{equation}
%
%
where $\{w_i^k\}_k, k=1,\ldots,K$ denote the $K$ noise samples, which are drawn from the noise distribution $p_n$, 
to train the model on the target word $w_i$. The gradient of $\mathcal{J}(\theta)$ is given by
%
\begin{align}
\label{eq:dllog}
  \frac{\partial\mathcal{J}(\theta)}{\partial\theta} &= -\sum_{w_i}^D \Big( p_{c_i}^{w_i}(0) \frac{\partial log(p_{\theta}^{c_i}(w_i))}{\partial\theta} \Big. \nonumber  \\
 &\quad  \Big. - \sum_{k}{p_{c_i}^{w_i^k}(1) \frac{\partial log(p_{\theta}^{c_i}(w_i^k))}{\partial\theta}}  \Big) 
\end{align}
%
NCE training of a neural network follows the standard back-propagation algorithm applied to the objective 
function~(\ref{eq:llog}) and its gradient~(\ref{eq:dllog}). 
More details about NCE and its gradient derivation can be found in~\cite{Gutmann2012}.

\subsection{NCE vs Importance Sampling}
\label{ssec:is}
The authors of~\cite{Rafal2016} have shown that NCE and Importance Sampling (IS) are closely related, with the main difference
is that NCE is defined as a binary classifier between samples drawn from data or noise distributions with a logistic loss, whereas
IS is a multi-class classifier, which uses \textit{softmax} and a cross-entropy loss. Hence, the authors concluded that 
IS is theoretically a better choice than NCE. 
The results reported, however, showed a minor difference in performance ($2.4$ points in perplexity). Moreover, training using IS
can be very difficult and requires a careful control of the samples variance, which can lead otherwise to unstable learning
as was reported in~\cite{Bengio2008}. Hence, an adaptive IS may use a large number of samples to solve this problem whereas
NCE is more stable and requires a fixed small number of noise samples (e.g., 100) to achieve a good performance~\cite{Mnih2012,Zoph2016}.
Furthermore, the network learns to \textit{self-normalize} during training using NCE. As a results, and on the contrary to IS, 
the \textit{softmax} is no longer required during evaluation, which makes NCE an attractive choice to train 
large vocabulary NNLM.
The next section will show how NCE can be efficiently implemented in batch mode training. 

\section{Batch Noise Contrastive Estimation}
\label{sec:bnce}
Although NCE is a good alternative to train large vocabulary LMs, it is not well-suited for batch mode training 
on GPUs. More particularly, each target word in the batch uses a different set of noise samples,
which makes it difficult to formulate the learning using dense matrix operations.  As a result, 
the training time significantly increases. To alleviate this problem, noise
samples can be shared across the batch~\cite{Zoph2016}. 

This paper proposes an extension of NCE to batch mode (B-NCE) training. This approach does not require any sampling
and can be formulated using dense matrix operations. Furthermore, we can show that this solution optimally 
approximates the sampling from a unigram distribution, which has been shown to be a good noise distribution choice \cite{Mnih2012,Zoph2016}.  

The main idea here is to restrict the vocabulary, at each forward-backward pass, to the 
target words in the batch (words to predict) and then replace the \textit{softmax} 
function by NCE. In particular, these words play alternatively the role of targets and noise samples. 
That is, for a target word $w_i$, at batch index $i$, the rest of the target batch (the remaining target 
words at the other batch indices $j$, $j\neq i$) are considered to be the noise samples. 
The rest of this section introduces the mathematical formulation of B-NCE to efficiently calculate the error with respect to
the output layer weights and biases, as well as the error at the previous layer in batch training, 
using the objective function~(\ref{eq:llog}) and its gradient~(\ref{eq:dllog}).
 
\subsection{LM Training using B-NCE}
\label{ssec:LMT}

Let $B$, $H$ and $V$ be the sizes of the batch, the last hidden layer and the vocabulary, respectively.
The matrix $L^t$ (size $B \times H$) will denote the evaluation of the last hidden layer at time $t$ on the current batch. 
Let $V^t=\{w_b^t\}_{b=1}^{B}$ be the target words in the batch at time $t$ and let $W$ (size $H \times V$) and $C$ ($1 \times V$) denote the 
hidden-to-output weight matrix and bias vector, respectively. Our goal here is to calculate the error (delta) of the output weights $W$ 
and biases $C$, as well as the error at the previous layer $L^t$.

The output layer evaluation in a feed-forward pass of B-NCE, at time $t$, is calculated by restricting the output layer to $V^t$. 
That is, we use the sub-matrix weights $W^t$ ($H \times B$) and sub-vector bias $C^t$ ($1\times B$), 
which are obtained by restricting $W$ and $C$ to $V^t$. Hence, the B-NCE network output $O^t$, at time $t$, is given by
\vspace{1mm}
\begin{equation}
\label{eq:bnce-out}
\displaystyle{ O^t = \frac{exp(L^t \cdot W^t \oplus C^t)}{Z} \quad (\text{size} \hspace{1mm} B \! \times \! B)}  
\vspace{1mm}
\end{equation}
$\oplus$ adds the vector $C^t$ to each row of the left-hand matrix. Note that (\ref{eq:bnce-out}) is the B-NCE matrix form of (\ref{eq:prob}) 
in batch training.

Now, let $N^t = p_n(\{w_b^t\})$ ($1\times B$) be the probability of the target words in the batch according to the noise distribution $p_n$. 
In order to evaluate the gradient of the objective function w.r.t. the output weights and biases, we first 
define the normalization matrix $Y^t$ according to 
\vspace{1mm}
\begin{equation}
\label{eq:norm}
\displaystyle{ Y^t = O^t \oplus (B-1) \cdot N^t \quad (\text{size} \hspace{1mm} B \! \times \! B)}  
\vspace{1mm}
\end{equation}
$Y^t$ is simply the normalization term in equations (\ref{eqn:true}) and (\ref{eqn:false}) with $K=B-1$. This is a direct result of using
the rest of the words in the output batch as NCE noise samples, for each target word $w_b^t$. In doing so, we eliminate the sampling step. 

In order to calculate~(\ref{eq:dllog}) w.r.t. the output weights and biases, 
we first introduce the auxiliary B-NCE gradient matrix $\mathcal{G}^t$ ($B\times B$) at time $t$, given by

\vspace{1mm}
$\begin{displaystyle}
  \mathcal{G}^t(i,j) = \left.
  \begin{cases}
   \frac{O_B^t(i,j)}{Y^t(i,j)} & \text{ if } i \neq j  \\
	\vspace{2mm}
   \frac{-(B-1)\cdot N^t(i)}{Y^t(i,j)} & \text{otherwise } 
  \end{cases}
  \right.
\end{displaystyle}$
\vspace{0mm}

%
$\mathcal{G}^t$ is nothing but the element-wise division of $O^t$ and $Y^t$, after replacing the diagonal of $O^t$ by $-(B-1)\cdot N^t$. 
Then, applying the NCE gradient derivation given by (\ref{eq:dllog}), B-NCE calculates the output weight error $\Delta W^t$, 
the output bias error $\Delta C^t$ as well as the error $E(L^t)$ at the previous layer according to
\vspace{-1mm}
\begin{eqnarray}
\label{eq:bnce-up}
 \Delta W^t &=& {L^t}^\top \cdot \mathcal{G}^t   \\
 \Delta C^t &=& \displaystyle{\sum_{row} \mathcal{G}^t} \\ 
   \label{eq:lastH} E(L^t)   &=& \mathcal{G}^t \cdot {W^t}^\top 
\vspace{-1mm}
\end{eqnarray}
Once the error $E(L^t)$ is propagated to the last hidden layer $L^t$ using (\ref{eq:lastH}), the learning of the rest of the network follows the standard back-propagation algorithm. 

After processing the complete training data, each word $w$ in the vocabulary will be used exactly $(B-1) \times count(w)$ as a noise sample.
This is strictly equivalent to sampling from a unigram noise distribution, which shows that B-NCE is an optimal implementation of NCE
using unigram as noise distribution. 
We should also mention that some words may occur more than once in the batch. This observation should be taken into consideration before 
updating the weights and biases. 

\subsection{Adaptive B-NCE}

The proposed B-NCE approach as defined above uses a fixed number of noise samples ($B-1$), which is dependent on the batch size.
In cases where the latter is small (e.g., $B \leq 100$), B-NCE can be extended to use an additional K noise samples. This can be done by simply
drawing an additional K samples form the noise distribution $p_n$, and share them across the batch as it was done in~\cite{Zoph2016}. 
The adaptive B-NCE follows the exact same steps described above using the extended output weight sub-matrix $W^t_{B+K}$ ($H \times (B+K)$), 
and the extended sub-vector bias $C^t_{B+K}$ ($1\times (B+K)$) to evaluate (\ref{eq:bnce-out}), whereas (\ref{eq:norm}) becomes
\vspace{-1mm}
\begin{equation}
\label{eq:extnorm}
 \displaystyle{ Y^t = O^t \oplus (B+K-1) \cdot N^t_{B+K} \quad \! \!\! \hspace{1mm} B \! \times \! (B+K) }  \!\!\! 
\vspace{1mm}
\end{equation}
where $N^t_{B+K}=\left[ N^t \hspace{0.2mm} , \hspace{0.2mm} N^t_K \right]$. $N^t_K$ are the 
probabilities of the additional $K$ noise samples using the noise distribution $p_n$.

\section{Experiments and Results}
\label{sec:exp}
%
%
To evaluate the proposed B-NCE approach, we conducted a set of LM experiments on two different corpora. 
Namely, the Large Text Compression Benchmark (LTCB)~\cite{Mahoney2011}, 
which is an extract of the enwik9 dataset, and the very large One Billion Word Benchmark (OBWB)~\cite{Chelba2013}.

%
The LTCB data split and processing is the same as the one used in~\cite{Oualil2016,Oualil2016b}. In particular,
the LTCB vocabulary is limited to the 80K most frequent words with all remaining words replaced by $<$unk$>$. 
Similarly to RNNLM toolkit~\cite{Mikolov2011}, we add a single $<$/s$>$ tag at the end of each sentence whereas the begin tag $<$s$>$ is not used.  
The resulting corpus size is $133M$ with an $<$unk$>$ rate of $1.43\%$ for the training set and $2.30\%$ for the test set. 
The second corpus is the OBWB, which contains $\approx0.8B$ tokens with a vocabulary size of $\approx 0.8M$ words. 
The data processing follows the description provided in~\cite{Chelba2013} leading to an $<$unk$>$ rate of 
$\approx0.3\%$. Similarly to LTCB, a single $<$/s$>$ tag is added at the end of each sentence. In the experiments described below, 
the first 5 held-out sets are used for validation whereas the remaining 45 sets are used for testing. 
The obtained results, however, showed that the models perplexity on these two sets is comparable, with an average difference of less than 0.5 points in perplexity. 

The primary motive of using LTCB, with its medium vocabulary size (80K), is to be able to compare the performance of LMs
trained using NCE to their counterparts that are trained using the full \textit{softmax}. 
When using NCE to train the models, the evaluation is either performed using the NCE constant $Z$ for 
normalization ($\text{PPL}^n$), in this case the target word probabilities are given by (\ref{eq:prob}), 
or using the \textit{softmax} function ($\text{PPL}^f$), which calculates these probabilities using (\ref{eq:softmax}).
The difference in performance between these metrics will evaluate the ability of the models to learn to self-normalize after training. 
For a comprehensive comparison of the different models, we also report the Number of Parameters (NoP) required by each model as well as its Training Speed (TS), 
which is calculated as the number of words processed per second (w/s) during training. 
All experiments were conducted on a \textbf{single Titan-X GPU}.

\subsection{Baseline Models}

In order to assess the gap among established NNLMs, 
this paper also presents a comparative study of different standard architectures with comparable NoPs. That is, we report results 
for the standard Feed-forward network (FFNN)~\cite{Bengio2003a}, the Recurrent Neural Network (RNN)~\cite{Mikolov2011} as well as 
the Long-Short Term Memory network (LSTM)~\cite{Sundermeyer12}. Our RNN implementation uses a projection weight matrix to decouple
the word embedding and the hidden layer sizes. We also report results after adding a bottleneck fully-connected ReLu layer 
right before the output layer in the recurrent models. These models are marked with the prefix \textit{ReLu} in the tables below. 
Each of the models is trained using the proposed B-NCE approach and the shared noise NCE (S-NCE)~\cite{Zoph2016}. 
For the LTCB corpus, we also report results of the models trained with the full \textit{softmax} function. This is the primary motive for using this corpus.  
We would like also to highlight that the goal of this paper is not about improving LMs performance but rather 
showing how a significant training speed-up can be achieved without compromising the models performance for large 
vocabulary LMs. Hence, we solely focus our experiments on NCE as a major approach to achieve this 
goal~\cite{Chen2015,Mnih2012,Zoph2016} in comparison to the reference full \textit{softmax} function. 
Comparison to other training approaches such as importance sampling will be conducted in future work. 

\subsection{LTCB Experiments}

For the LTCB experiments, the embedding size is fixed at 200, the 5-gram FFNN has two hidden layers, whereas 
RNN and LSTM use a single recurrent layer. All non-recurrent layers use ReLu as activation function. 
More details about the models architectures are shown in Table~\ref{tab:ltcbconfig}, where ``(R)'' 
stands for recurrent and ``(B)'' for bottleneck.
The batch size is fixed at 400 and the initial learning rate is set to 0.4. The latter is halved when no improvement on 
the validation data is observed for an additional 7 epochs. 
We also use a norm-based gradient clipping with a threshold of 5 but we do not use dropout. Moreover, 
B-NCE and S-NCE use the unigram as noise distribution $p_n$. 
Following the setup proposed in~\cite{Mnih2012,Zoph2016}, S-NCE uses $K=100$ noise samples, whereas B-NCE 
uses only the target words in the batch (K=0). 
Note that S-NCE will process and update $B+K$ words at its output layer during each forward-backward pass, 
whereas B-NCE updates only $B$ words. Similarly to~\cite{Chen2015}, the NCE normalization constant is set to $Z=exp(9)$, 
which approximates the mean of the normalization term using \textit{softmax}.

\vspace{-3mm}
\renewcommand{\tabcolsep}{2pt}
\begin{table}[!h]
    \caption{\label{tab:ltcbconfig} Model architecture for LTCB experiments.}
    	\vspace{-1.5mm}
  \centerline{
  \begin{tabular}{| c || c |}
     \hline
       Model &  Architecture \\
      \hline
      5-gram FFNN   &  4$\times$200$-$600$-$400(B)$-$V  \\ 
      RNN           &  200$-$600(R)$-$V  \\ 
      ReLu-RNN      &  200$-$600(R)$-$400(B)$-$V \\
      LSTM          &  200$-$600(R)$-$V \\
      ReLu-LSTM     &  200$-$600(R)$-$400(B)$-$V \\ 
      \hline
  \end{tabular} }
  \vspace{-3mm}
\end{table}

The LTCB results reported in Table~\ref{tab:ltcb} clearly show that B-NCE reduces the training time by a factor 
of 4 to 8 with a slight degradation in the models performance compared to \textit{softmax}. 
Moreover, we can also see that B-NCE slightly outperforms S-NCE while being faster and simpler
to implement. In particular, B-NCE does not require the sampling step since it uses the rest of the output words in the batch itself
as noise samples to train the model on each target word. This can be efficiently implemented using dense matrix operations (see Sections \ref{sec:bnce}). 
Table~\ref{tab:ltcb} also shows that PPL$^n$ is close from PPL$^f$, which typically reflects 
that the models trained using NCE are able to self-normalize, where the normalization term using softmax is, in average, very
close from the NCE constant Z. We have also observed in our experiments that the models degradation and the gap between PPL$^f$ and PPL$^n$ strongly depend on the 
amount of training data, the vocabulary size as well as the size of the last hidden layer. More particularly, increasing the training data
leads to a more stable learning and therefore to a smaller gap between these two metrics and a much lower degradation of the models performance (see OBWB experiments below). 
\renewcommand{\tabcolsep}{2pt}
\begin{table}[!h]
  \caption{\label{tab:ltcb} LMs performance on LTCB.}
  \vspace{-2mm}
  \centerline{
  \begin{tabular}{| c || B | B | C | C |}
     \hline
                                 & $\text{PPL}^n$ & $\text{PPL}^f$  & TS (w/s) & NoP  \\
      \hline
      5-gram FFNN (\textit{softmax}) &  ---   &  110.2  &  8.4K  & 48.8M  \\ 
      5-gram FFNN (S-NCE)            & 129.8  &  125.4  & 29.1K  & 48.8M  \\   
      5-gram FFNN (B-NCE)            & 119.4  &  113.7  & 35.1K  & 48.8M  \\ 
      \hline
      \hline
      RNN(\textit{softmax})       &  ---   &  79.7   &  5.9K  & 64.6M  \\ 
      RNN (S-NCE)                 &  88.7  &  84.2   & 37.8K  & 64.6M  \\ 
      RNN (B-NCE)                 &  87.6  &  82.5   & 43.7K  & 64.6M  \\ 
      \hline
      ReLu-RNN (\textit{softmax}) &  ---   &  69.5   &  8.6K  & 48.8M  \\ 
      ReLu-RNN (S-NCE)            &  80.2  &  77.3   & 30.9K  & 48.8M  \\ 
      ReLu-RNN (B-NCE)            &  79.4  &  76.0   & 36.7K  & 48.8M  \\ 
      \hline
      \hline
      LSTM (\textit{softmax})     &  ---   &  62.5   &  8.9K  & 66.0M  \\ 
      LSTM (S-NCE)                &  77.4  &  73.1   & 27.2K  & 66.0M  \\ 
      LSTM (B-NCE)                &  70.9  &  68.3   & 37.1K  & 66.0M  \\ 
      \hline
      ReLu-LSTM (\textit{softmax})&   ---  &  59.2   &  8.2K  & 50.3M   \\ 
      ReLu-LSTM (S-NCE)           &  68.4  &  67.1   & 26.9K  & 50.3M   \\ 
      ReLu-LSTM (B-NCE)           &  64.9  &  62.4   & 32.0K  & 50.3M   \\ 
      \hline
  \end{tabular} }
	\vspace{-1mm}
\end{table}
\vspace{-2mm}

We can also conclude from Table~\ref{tab:ltcb} that the additional ReLu layer improves the performance while
significantly decreasing the number of parameters (NoP). This conclusion is valid for both, RNN and LSTM. 
These results confirm that adding a fully-connected bottleneck layer can significantly boost the performance
of recurrent models. This idea has been previously used in computer vision tasks in combination 
with Convolutional Neural Networks (CNN)~\cite{Simonyan2014}, as well as in speech recognition~\cite{Hasim2014}, 
where the fully-connected layer is used as pat of the LSTM recurrent module.
\subsection{One Billion Word Benchmark Experiments}
The OBWB experiments are similar to LTCB with minor differences. Namely, the embedding size is set to 500 for all models, 
the batch size is fixed at 500, S-NCE uses $K=200$ noise samples and the initial learning rate is set to 1.0. 
Given that the vocabulary size is $\approx 0.8M$, it was not possible to train the language models using the full \textit{softmax} function. 
Therefore, we only report results for B-NCE and S-NCE. More details about the models configuration are shown in Table~\ref{tab:obwconfig}. 
\vspace{-2mm}
\renewcommand{\tabcolsep}{2pt}
\begin{table}[!h]
    \caption{\label{tab:obwconfig} Model architecture for OBWB experiments.}
		  \vspace{-2mm}
  \centerline{
  \begin{tabular}{| c || c |}
     \hline
       Model &  Architecture \\
      \hline
      5-gram FFNN   &  4$\times$500$-$1500$-$600(B)$-$V  \\ 
      RNN           &  500$-$1500(R)$-$V  \\ 
      ReLu-RNN      &  500$-$1500(R)$-$600(B)$-$V \\
      LSTM          &  500$-$1500(R)$-$V \\
      ReLu-LSTM     &  500$-$1500(R)$-$600(B)$-$V \\ 
      \hline
  \end{tabular} }
  \vspace{-1.5mm}
\end{table}
\vspace{-2mm}

The OBWB results in Table~\ref{tab:obw} generally confirm the LTCB conclusions. That is, 
B-NCE slightly outperforms S-NCE while being faster and simpler to train (training speed in $3^{rd}$ column). Moreover, these results 
also show a much smaller difference between PPL$^f$ and PPL$^n$ compared to LTCB, which suggests that the models learned to better 
self-normalize due to the larger amount of training data. Similarly to LTCB, we can see that the additional ReLu
helps reducing the NoPs while improving or maintaining the models performance for RNN and LSTM. 
\vspace{-3mm}
\renewcommand{\tabcolsep}{2pt}
\begin{table}[!h]
  \caption{\label{tab:obw} LMs performance on OBWB.}
	\vspace{-2mm}
  \centerline{
  \begin{tabular}{| c || B | B | C | C |}
     \hline
                                 & $\text{PPL}^n$ & $\text{PPL}^f$  & TS (w/s) & NoP  \\
      \hline
      5-gram FFNN (S-NCE)        &  86.3  &   84.4  & 12.3K  & 0.88B  \\   
      5-gram FFNN (B-NCE)        &  81.9  &   80.6  & 13.4K  & 0.88B  \\ 
      \hline
      \hline
      RNN (S-NCE)                &  70.8   & 67.6   & 24.9K  & 1.59B  \\ 
      RNN (B-NCE)                &  63.4  &  61.4   & 27.8K  & 1.59B  \\ 
      \hline
      ReLu-RNN (S-NCE)           &  59.6  &  59.1   & 20.1K  & 0.88B  \\ 
      ReLu-RNN (B-NCE)           &  56.9  &  56.6   & 23.1K  & 0.88B  \\ 
      \hline
      \hline
      LSTM (S-NCE)               &  51.3  &  50.3   & 12.0K  & 1.60B  \\ 
      LSTM (B-NCE)               &  48.6  &  48.1   & 13.1K  & 1.60B  \\ 
      \hline
      ReLu-LSTM (S-NCE)          &  51.0  &  50.9   & 13.0K  & 0.89B   \\ 
      ReLu-LSTM (B-NCE)          &  49.2  &  48.8   & 14.7K  & 0.89B   \\ 
      \hline
  \end{tabular} }
	\vspace{-1.5mm}
\end{table}
\vspace{-1mm}

In comparison to other results on the OBWB. We can see that the small ReLu-RNN achieves a close performance from the very large
RNN model (PPL = 51.3 and NoP = 20B) proposed in~\cite{Chelba2013}. Moreover, the performance of the small ReLu-LSTM is comparable to the 
LSTM models proposed in~\cite{Zoph2016} and~\cite{Rafal2016} which use large hidden layers. In particular, the first paper 
trains a large 4-layers LSTM model using S-NCE on 4 GPUs (PPL = 43.2 and NoP = 3.4B),
whereas the second uses a recurrent bottleneck layer~\cite{Hasim2014} and a total of $K=8192$ noise samples with importance 
sampling on 32 Tesla K40 GPUs. 

\section{Conclusions and Future Work}
\label{sec:cc}
We have presented a batch-NCE approach which
allows a fast and simple training of large vocabulary LMs. 
This approach eliminates the sampling step required in standard NCE
and can be fully formulated using dense matrix operations.
Experiments on LTCB and OBWB have shown that this approach 
achieves a comparable performance to a \textit{softmax} 
function while significantly speeding-up the training.
While the evaluation focused on NCE performance, future experiments 
will be conducted to evaluate B-NCE in comparison to other alternatives of \textit{softmax}. 

\section{Acknowledgment}
\label{sec:ack}
This research was funded by the German Research Foundation (DFG) as part of SFB 1102.

\bibliographystyle{IEEEtran}

\bibliography{mybib}

\end{document}